\documentclass{article}
\usepackage{spconf,amsmath,graphicx}
\usepackage{caption, subcaption}
\graphicspath{{new_report_data}}

\title{Pyramid Fusion Dark Channel Prior for Single Image Dehazing}
\name{Qiyuan Liang$^{\star}$ \qquad Bin Zhu$^{\dagger}$ \qquad Chong-Wah Ngo$^{\ddagger}$}
  
\address{$^{\star}$ École polytechnique fédérale de Lausanne, Switzerland \\
      $^{\dagger}$ City University of Hong Kong, Hong Kong \\
      $^{\ddagger}$ Singapore Management University, Singapore}
\begin{document}
%\ninept
%
\maketitle
\begin{abstract}
In this paper, we propose the pyramid fusion dark channel prior (PF-DCP) for single image dehazing. Based on the well-known Dark Channel Prior (DCP), we introduce an easy yet effective approach PF-DCP by employing the DCP algorithm at a pyramid of multi-scale images to alleviate the problem of patch size selection. In this case, we obtain the final transmission map by fusing transmission maps at each level to recover a high-quality haze-free image. Experiments on RESIDE SOTS show that PF-DCP not only outperforms the traditional prior-based methods with a large margin but also achieves comparable or even better results of state-of-art deep learning approaches. Furthermore, the visual quality is also greatly improved with much fewer color distortions and halo artifacts.
\end{abstract}
\begin{keywords}
Traditional method, single image dehazing, image pyramid, dark channel prior
\end{keywords}
\section{Introduction}
\label{sec:intro}

Haze is ubiquitous in life, which comes from fog, sandstorms, or pollutants in the air. Haze in the air largely degrades the quality of the photo. Many details of the scene become blurry and the visibility is reduced. From the aspect of aesthetic, the images without haze are much clearer and visually pleasing than the hazy ones. For these reasons, haze removal has been studied for decades, yet still a challenging task in computer vision.

The key problem of haze removal is the strategy to dehaze an image. Hautiére et al. \cite{hautiere2006automatic} proposed an equation to model hazy images captured by the camera:
\begin{equation}
    \label{eqn:haze model}
    I(x) = J(x)t(x) + A(1-t(x)),
\end{equation}
where $I$ is the observed hazy image, $J$ is the real scene radiance, $A$ is the atmospheric light, $t$ is the transmission rate. The goal of the image dehazing is to recover haze-free scene $J$, with only the knowledge of $I$. If we are able to estimate $A$ and $t$, we can recover the real scene $J$ as follows:
\begin{equation}
    \label{eqn:dehaze}
    J(x) = \frac{I(x)-A}{t(x)}+A.
\end{equation}

However, $A$ and $t$ are both unknown, and it is impossible to estimate their values from a single hazy image without any constraint. Early approaches developed a strong prior or assumption of the difference between the haze-free and hazy images. Tan \cite{tan2008visibility} maximized  the local contrast to restore the clear images given the fact that clear images have higher contrast compared to hazy ones. He et al. \cite{he2010single} utilized the prior knowledge that clear images have a dark channel whereas hazy images do not. Meng et al. \cite{meng2013efficient} introduced a boundary constraint to help formulate the transmission map. \cite{zhu2015fast} proposed a color attenuation prior (CAP), which models a linear relationship between the scene depth, brightness, and saturation. Berman et al. \cite{berman2016non} suggested a non-local prior (NLP), using only 500 distinct colors to quantize the image and clustering different pixels according to their colors into haze-lines for dehazing.

In the emerging years of deep learning, researchers focus on building models to generate the transmission map, $t(x)$, and atmospheric light, $A$, separately. Cai et al. \cite{cai2016dehazenet} used a convolutional neural network (CNN) to estimate the transmission rate and use the atmospheric scattering model to dehaze the image. Ren et al. \cite{ren2016single} introduced a multi-scale CNN model to improve the quality of the transmission map. However, estimating two parameters individually might cause larger errors. Li et al. \cite{li2017aod} unified the transmission map and the atmospheric light into one parameter and use it to estimate the clear image. Later, Ren et al. \cite{ren2018gated} designed a coarse-to-fine fusion network, which recovers the clear image directly from the hazy one. Deep learning has gained huge success by surpassing the physical models built by humans and exploring the tremendous power of computation. Nevertheless, it is troublesome to train a deep learning model so that it performs well. Not only a large labeled dataset is required, but also the training process is time-consuming, let alone the problems of interpretability and robustness \cite{marcus2018deep}, which is required to be employed for productization. 

Among all traditional algorithms, due to the effectiveness and robustness, DCP serves as a building block to many haze-removal tasks \cite{xu2012fast,pan2016blind}. Besides, DCP is also broadly used as a standard baseline of the performance evaluation in recent works using deep learning \cite{cai2016dehazenet, li2017aod,  ren2018gated, qu2019enhanced, zhang2018densely, dong2020fd}. However, DCP still suffers from the problem of patch size selection. Inappropriate patch size would introduce an inaccurate estimation of the global atmospheric light and a poor transmission map \cite{lee2016review}, which leads to dis-satisfactory haze removal results.

In this paper, we propose the pyramid fusion dark channel prior (PF-DCP) using an iterative weighted fusion strategy to address the problem of patch size selection. Through incorporating multi-scale information of the transmission map, we demonstrate that the proposed method outperforms DCP by a large margin with a similar computational cost. The halo artifacts of the original DCP are greatly reduced, and the dehazed images become much smoother in colors and visually pleasing. Although the idea of using multi-scale information is well-established in various computer vision tasks, its application as a traditional algorithm in this field has not been proposed yet. The novelty of this paper lies in two aspects by solving two critical problems. First, there is no evaluation metric to select the best patch size. Thus, we adopt pyramid fusion among different sizes to alleviate the problem. Second, the process of transmission map refinement helps maintain a similar level of sharpness at the final dehazed images, which prevents the algorithm from degrading the image quality. Furthermore, as a traditional algorithm that is interpretable and robust, PF-DCP manages to produce superior results compared to other prior-based traditional algorithms and commensurable to the latest deep learning methods.

\section{Method}
\label{sec:method}

\subsection{Preliminaries}
\label{subsec: pre}
He et al. \cite{he2010single} proposed the Dark Channel Prior (DCP) for removing haze in one single image. It is observed that for non-sky outdoor areas, the intensity value of the dark channels of a haze-free image is close to zero. Dark channels are defined as the minimal value of three color channels of all pixels within a local patch, which can be defined as follows:
\begin{equation}
    \label{eqn:dark channel}
    J^{dark}(x) = \min_{y \in \Omega (x)} 
        \left(
            \min_{c \in \{r, g, b\}}
            J^c(y)
        \right),
\end{equation}
where $J^c$ is one color channel of $J$, $\{r,g,b\}$ is the three color channels, $\Omega(x)$ is the local patch centered at pixel $x$. The statistics of the haze-free outdoor images suggests that
\begin{equation}
    \label{eqn:dark channel zero}
    J^{dark}(x) \approx 0.
\end{equation}

\begin{figure}[h]
     \centering
     \begin{subfigure}[b]{0.07\textwidth}
         \centering
         \includegraphics[width=\textwidth]{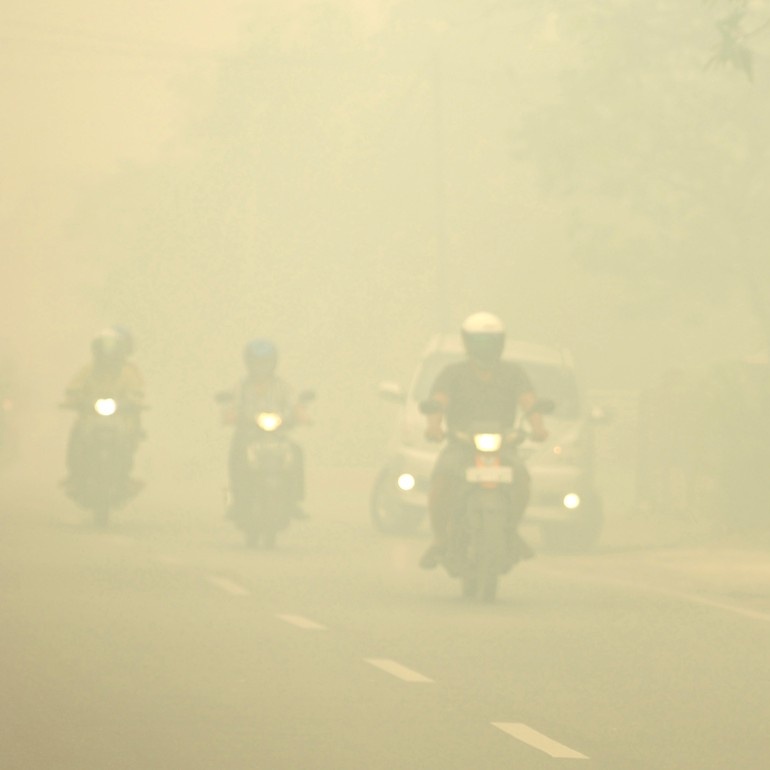}
     \end{subfigure}
     \begin{subfigure}[b]{0.07\textwidth}
         \centering
         \includegraphics[width=\textwidth]{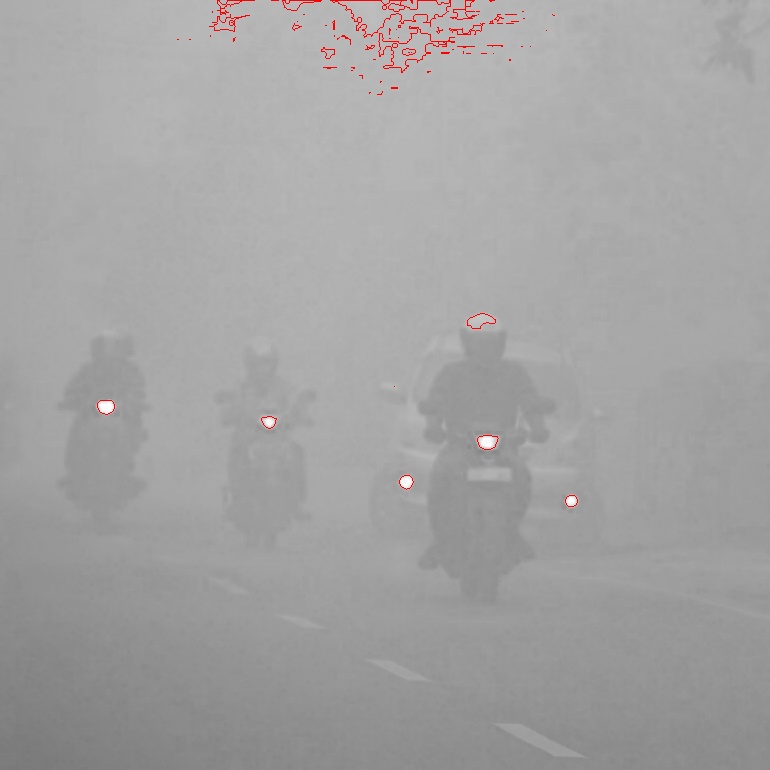}
     \end{subfigure}
     \begin{subfigure}[b]{0.07\textwidth}
         \centering
         \includegraphics[width=\textwidth]{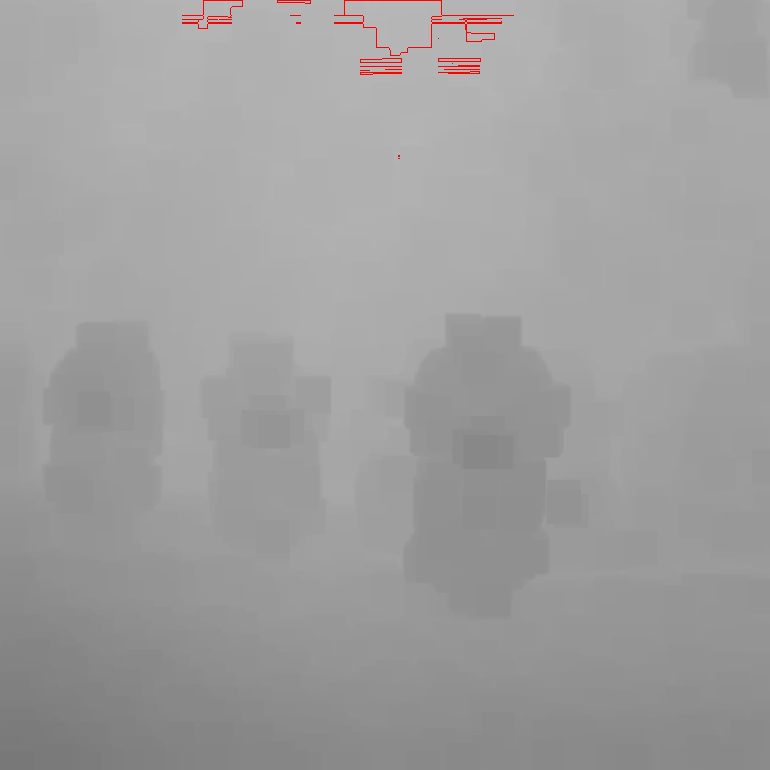}
     \end{subfigure}
     \hfill
     \begin{subfigure}[b]{0.07\textwidth}
         \centering
         \includegraphics[width=\textwidth]{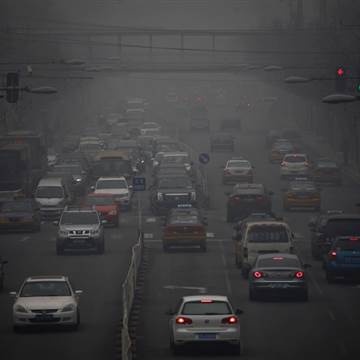}
     \end{subfigure}
     \begin{subfigure}[b]{0.07\textwidth}
         \centering
         \includegraphics[width=\textwidth]{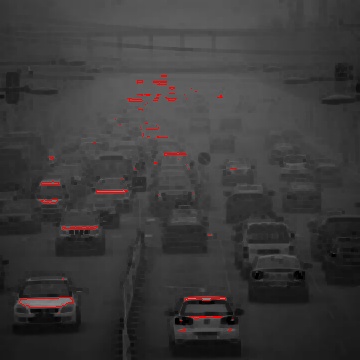}
     \end{subfigure}
     \begin{subfigure}[b]{0.07\textwidth}
         \centering
         \includegraphics[width=\textwidth]{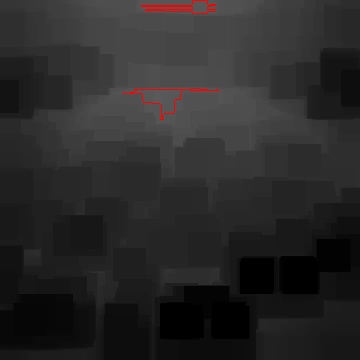}
     \end{subfigure}
        \caption{Areas used to estimate the global atmospheric light, percentage=1\%. From left to right are the hazy image, areas using local patch size $3\times3$ and local patch size $32\times32$ respectively.}
        \label{fig:dcp atom}
\end{figure}

\noindent \textbf{DCP Atmospheric Light Estimation.}
In DCP, the global atmospheric light is estimated from the most haze opaque areas, which are the pixels with the highest dark channel value. More specifically, it ranks the dark channel pixels by intensity values from high to low, then picks the top $0.1\%$ pixels. For these pixels, the one with the highest intensity value in the original hazy image $I$ is used as the estimation of atmospheric light. As shown in Fig. \ref{fig:dcp atom}, the patch size will influence the final atmospheric light estimation dramatically. A small patch size will make the process more subject to noise. On the contrary, a large patch size will lead to inaccurate estimation.

\begin{figure}[h]
     \centering
     \begin{subfigure}[b]{0.07\textwidth}
         \centering
         \includegraphics[width=\textwidth]{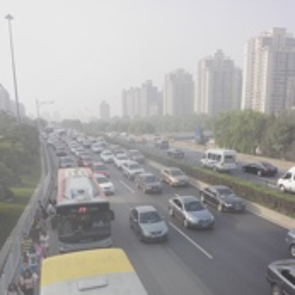}
     \end{subfigure}
     \begin{subfigure}[b]{0.07\textwidth}
         \centering
         \includegraphics[width=\textwidth]{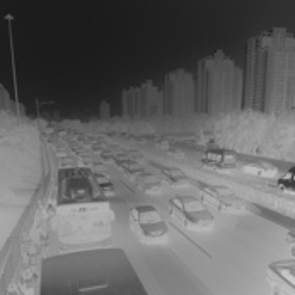}
     \end{subfigure}
     \begin{subfigure}[b]{0.07\textwidth}
         \centering
         \includegraphics[width=\textwidth]{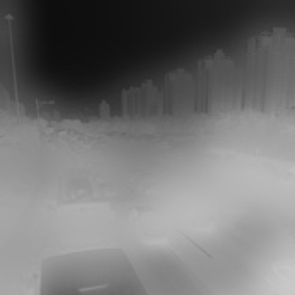}
     \end{subfigure}
     \hfill
     \begin{subfigure}[b]{0.07\textwidth}
         \centering
         \includegraphics[width=\textwidth]{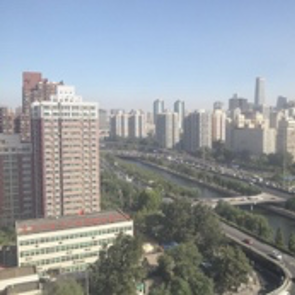}
     \end{subfigure}
     \begin{subfigure}[b]{0.07\textwidth}
         \centering
         \includegraphics[width=\textwidth]{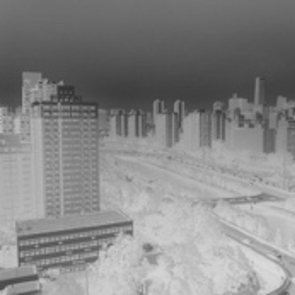}
     \end{subfigure}
     \begin{subfigure}[b]{0.07\textwidth}
         \centering
         \includegraphics[width=\textwidth]{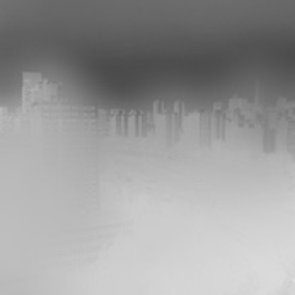}
     \end{subfigure}
        \caption{Transmission map estimation. From left to right are the hazy image, refined transmission map estimation using local patch size $3\times3$ and local patch size $32\times32$ respectively.}
        \label{fig:dcp trans}
\end{figure}

\noindent \textbf{DCP Transmission Map Estimation.}
Despite the fairly good dehazing performance of DCP, the same problem of patch size selection holds in the computation of the transmission map. As shown in Fig. \ref{fig:dcp trans}, details of hazy images can be preserved in the refined transmission map when a small patch size is applied, but it also introduces undesirable noise. When a large patch size is used, the transmission map becomes smoother, while the sharpness is reduced. Therefore, an appropriate patch size is also of great importance for a high-quality estimation of the transmission map.

\subsection{Pyramid Fusion Dark Channel Prior (PF-DCP)}
\label{subsec: pf-dcp}
In Section \ref{subsec: pre}, we introduce the basic ideas of DCP. The problems related to the patch size are exposed yet not solved. To address this issue, we propose the pyramid fusion dark channel prior (PF-DCP) based on the original DCP.

\begin{figure}[h]
    \centering
    \includegraphics[width=0.42\textwidth]{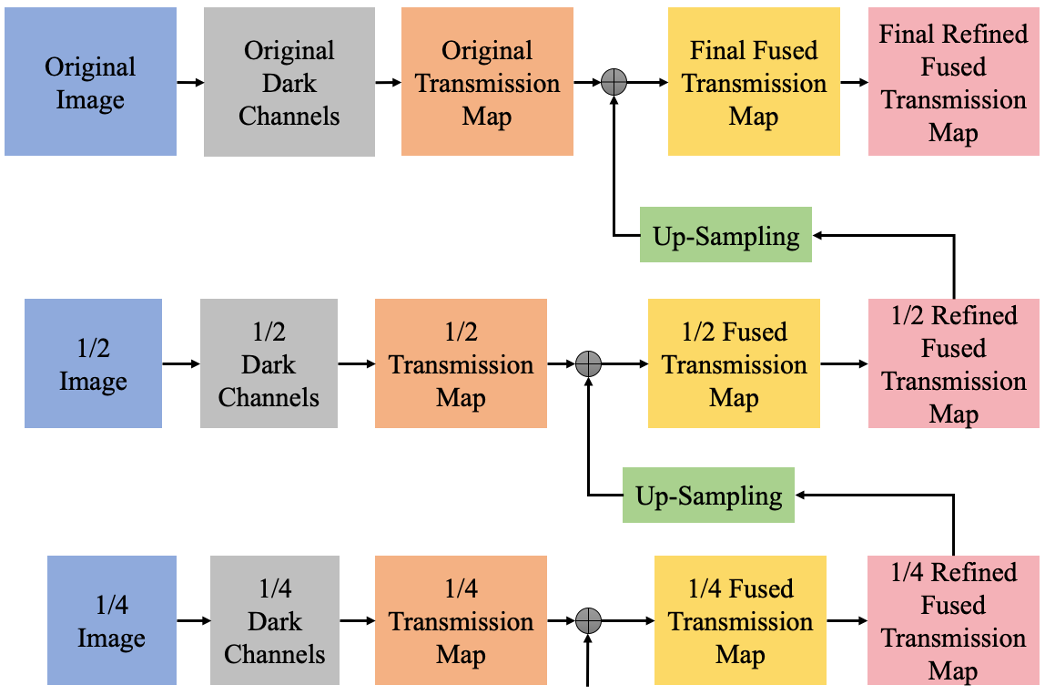}
    \caption{The architecture of Pyramid Fusion Dark Channel Prior.}
    \label{fig:pf-dcp_archi}
\end{figure}

The key steps of PF-DCP are shown in Fig. \ref{fig:pf-dcp_archi}. Note that we only show three levels of the image pyramid, in practice, we generate more levels of low-resolution images as long as their width and height are both above the patch size. At each level, the hazy image performs steps similar to DCP, including computing dark channels, estimating atmospheric light, estimating and refining transmission maps. The above procedures are similar to those of DCP. Later, after obtaining the transmission map at each level the lower-resolution transmission map is firstly up-sampled to the same size as the higher-resolution one, then we perform the weighted fusion among them. The process is conducted iteratively until the highest resolution, i.e., the original image size. The final transmission map will be further refined to recover the haze-free image.

\noindent \textbf{Image Pyramid Generation.} We generate a multi-level image pyramid starting from the original hazy image. We down-sample the image to half size until the size is less than the patch size. The level of the pyramid depends on the original size. In practice, it is normally within ten levels. After the refined transmission map estimation is computed at each level, it is up-sampled and restored iteratively to the original size. The transmission map with the original size is then used to recover the haze-free image.

\noindent \textbf{PF-DCP Atmospheric Light Estimation.} To address the atmospheric light estimation problem of the original DCP algorithm, we estimate it using dark channels from all pyramid levels, before the computation of the transmission maps. In this paper, we choose the maximal estimation value among all levels as the final atmospheric light estimation.

\noindent \textbf{PF-DCP Transmission Map Fusion.} Due to the absence of the evaluation metrics without ground truth, it is infeasible to perform automatic patch size selection. Therefore, we apply a fusion-based algorithm to incorporate multi-scale information. In this paper, we propose to perform an iterative weighted fusion of the transmission maps at each level and dehaze the original image via the final transmission map. In this way, the details of the original images are kept as much as possible while the final transmission map preserves multi-scale information. The reasons that we adopt a weighted fusion strategy between the transmission map lie in three aspects:
\begin{itemize}
    \item The up-sampled transmission map from lower resolution is structurally similar to the transmission map computed in higher resolution. Thus, performing a fusion between them is practical and technically sound.
    \item Different weight values offer much flexibility. If a smaller weight is set for the transmission map from the lower resolution, the final transmission map will contain more details of the scene. On the contrary, if a higher weight is set, the final transmission map might lose some details, but noises are reduced. Meanwhile, at least, part of the contribution of the final transmission map comes from the highest resolution, so details are preserved to some extent.
    \item The weight fusion is computationally efficient and can be an alternative way of different patch size selection. Specifically, a smaller patch size maintains more details and a larger patch size reduces noise in the transmission map. However, a larger patch size means a much slower computation speed. On the contrary, changing the weight has nearly no impact on the execution time, and has an even better performance compared to changing the patch size.
\end{itemize}

\noindent \textbf{Fused Transmission Map Refinement.} To further improve the quality of the results, we pay extra attention to the sharpness of the transmission map. Inspired by DCP, we apply guided image filtering \cite{he2010guided} to further refine the fused transmission map. The guided filter assumes a local linear relationship between the guiding image, which is the hazy image, and the output, which is the refined transmission map. Utilizing the structural information of the hazy image, the artifacts during the process of up-sampling and fusion are greatly reduced.

\noindent \textbf{Haze-free Image Recovery.} After obtaining the final transmission maps of the hazy images, we are able to easily recover the haze-free images using Eq. \ref{eqn:dehaze}. However, for the most haze opaque areas where the transmission $t$ is extremely small, dividing $I(x)-A$ by $t$ is prone to be influenced by noise \cite{he2010single}. Thus, we adopt a similar strategy to DCP to include a lower bound $t_0$ to make the algorithm more robust. The final formula to recover $J$ is
\begin{equation}
    \label{eqn:final dehaze}
    J(x) = \frac{I(x)-A}{\max(t(x), t_0)} + A.
\end{equation}

\section{Experiments}
\label{sec:experiments}

\subsection{Experimental Settings}
By default, the patch size is $15\times15$ for all pyramid levels. Top $0.1\%$ haze opaque pixels are used for global atmospheric light estimation and the lower bound of the transmission is $0.1$. The guided filter \cite{he2010guided} is adopted for transmission map refinement. To generate the image pyramid, we discard half of the rows and columns directly. For the image up-sampling, we use the nearest-neighbor interpolation. In our in-house experiments, there is no obvious performance difference by using other interpolation algorithms, such as bilinear, bicubic and nearest-neighbor. Using the extra training set of RESIDE \cite{li2017reside}, the best fusion weight of two transmission maps for every consecutive pairs, the lower and higher resolution level, is $4:1$ for indoor scenes $80:1$ for outdoor scenes. Compared to the original DCP, the computation cost of PF-DCP increases by $30\%$ on average.

For the parameter selection, we use images from Indoor Training Set (ITS) and Outdoor Training Set (OTS) of RESIDE \cite{li2017reside}. For testing, we evaluate the performance of the algorithm on the Synthetic Objective Testing Set (SOTS) of RESIDE, which contains 500 labeled indoor hazy images and 500 outdoor ones, and OHaze \cite{ancuti2018haze} for real hazy images. We use two commonly-used metrics for evaluation in single image dehazing: Peak Signal-to-Noise Ratio (PSNR) and Structural Similarity (SSIM).

\subsection{Experimental Results}
\noindent \textbf{Fusion Weight.} Since we have already generated an image pyramid and calculated transmission map at each level in PF-DCP, setting a higher weight value at the desired level of the image pyramid shall achieve a similar effect of selecting the desired patch size. Through experiments, we find out that fixating the patch size and changing the fusion weight achieves better performance both in speed and in quality than fixating fusion weight and changing patch size.

\begin{figure}[h]
    \centering
    \includegraphics[width=0.4\textwidth]{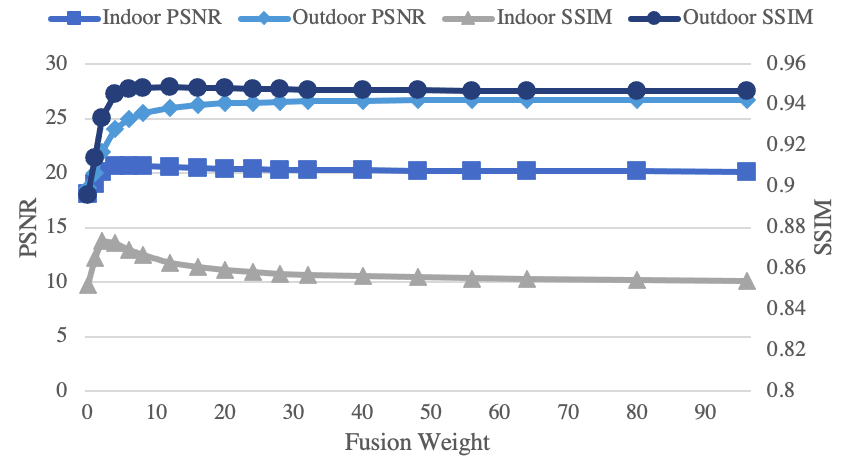}
    \caption{Results of different fusion weights on RESIDE ITS and OTS.}
    \label{fig:pf-dcp_fusion_ratio}
\end{figure}

For simplification, we fix the weight value between every two levels in the image pyramid. Fig. \ref{fig:pf-dcp_fusion_ratio} shows the results of using different fusion weights. The performance of outdoor scenes is much better than that of indoor ones. It is not surprising because DCP is for outdoor scenes in theory. For indoor scenes, they are sensitive to the change of fusion weights and achieves the best performance at a relatively small weight value. For outdoor scenes, they are the opposite. The plausible reasons are that there are more details and the relative depth change is large in a local patch of the indoor scenes.

\noindent \textbf{Comparison with state-of-the-art methods.} The proposed PF-DCP is compared against traditional prior-based approaches, DCP \cite{he2010single}, BCCR \cite{meng2013efficient}, CAP \cite{zhu2015fast}, and NLD \cite{berman2016non}. \cite{li2017reside}.

\begin{figure}[h]
     \centering
     \begin{subfigure}[b]{0.06\textwidth}
         \centering
         \includegraphics[width=\textwidth]{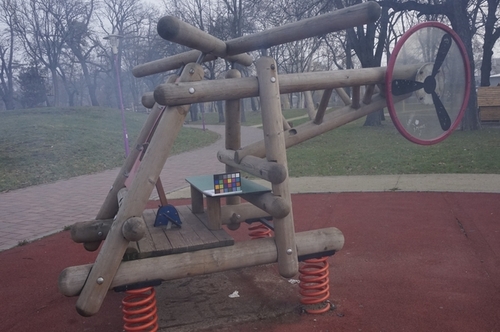}
     \end{subfigure}
     \begin{subfigure}[b]{0.06\textwidth}
         \centering
         \includegraphics[width=\textwidth]{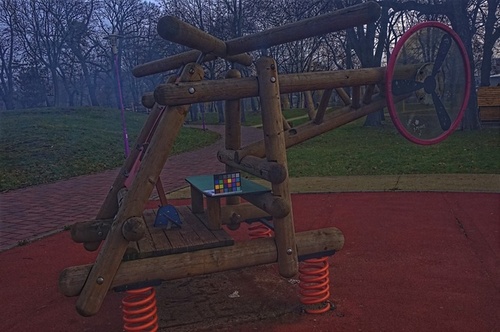}
     \end{subfigure}
     \begin{subfigure}[b]{0.06\textwidth}
         \centering
         \includegraphics[width=\textwidth]{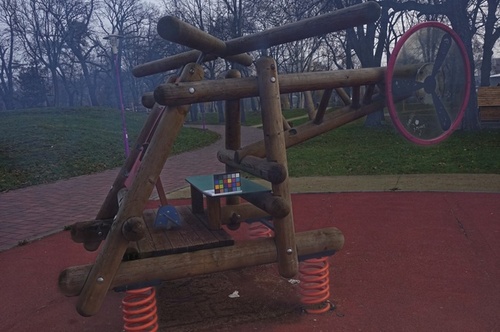}
     \end{subfigure}
     \begin{subfigure}[b]{0.06\textwidth}
         \centering
         \includegraphics[width=\textwidth]{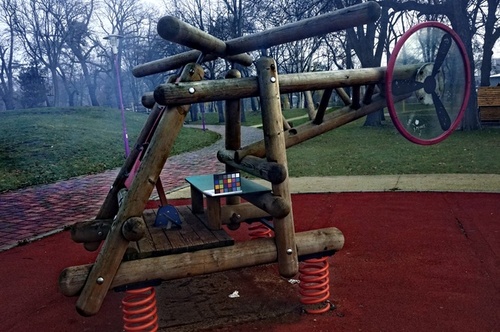}
     \end{subfigure}
     \begin{subfigure}[b]{0.06\textwidth}
         \centering
         \includegraphics[width=\textwidth]{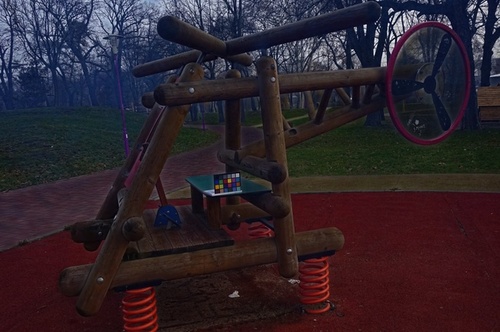}
     \end{subfigure}
     \begin{subfigure}[b]{0.06\textwidth}
         \centering
         \includegraphics[width=\textwidth]{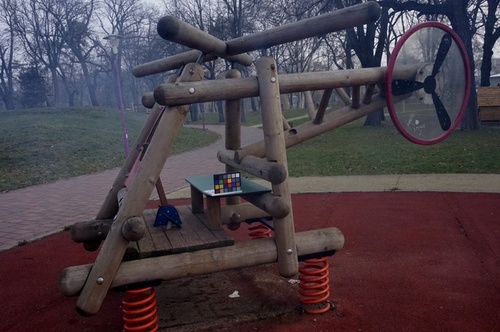}
     \end{subfigure}
     \begin{subfigure}[b]{0.06\textwidth}
         \centering
         \includegraphics[width=\textwidth]{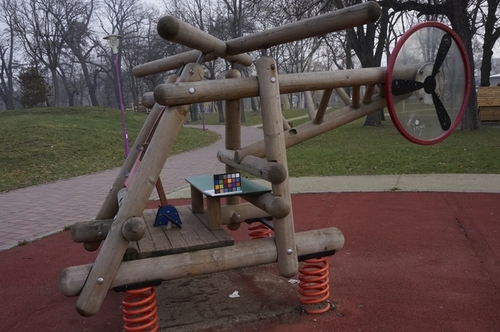}
     \end{subfigure}
          \begin{subfigure}[b]{0.06\textwidth}
         \centering
         \includegraphics[width=\textwidth]{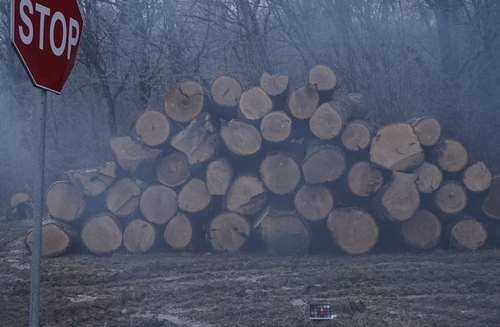}
     \end{subfigure}
     \begin{subfigure}[b]{0.06\textwidth}
         \centering
         \includegraphics[width=\textwidth]{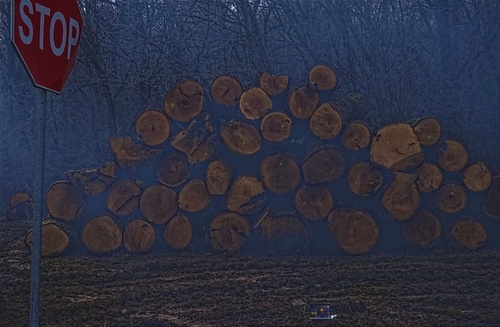}
     \end{subfigure}
     \begin{subfigure}[b]{0.06\textwidth}
         \centering
         \includegraphics[width=\textwidth]{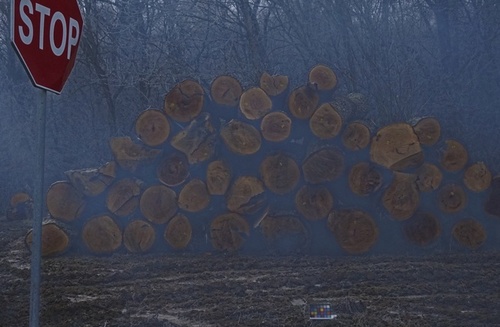}
     \end{subfigure}
     \begin{subfigure}[b]{0.06\textwidth}
         \centering
         \includegraphics[width=\textwidth]{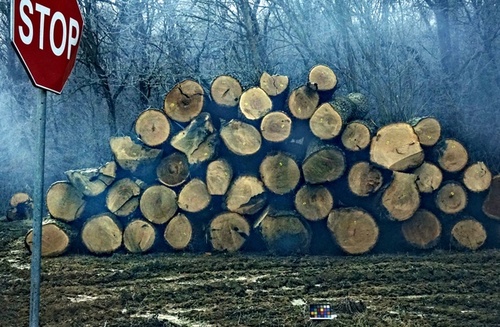}
     \end{subfigure}
     \begin{subfigure}[b]{0.06\textwidth}
         \centering
         \includegraphics[width=\textwidth]{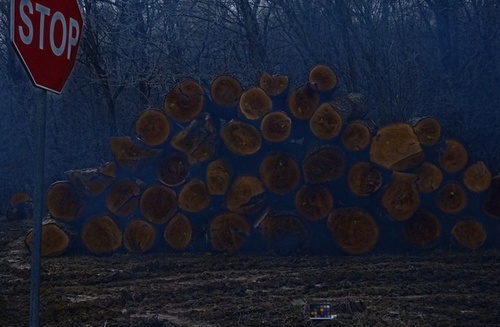}
     \end{subfigure}
     \begin{subfigure}[b]{0.06\textwidth}
         \centering
         \includegraphics[width=\textwidth]{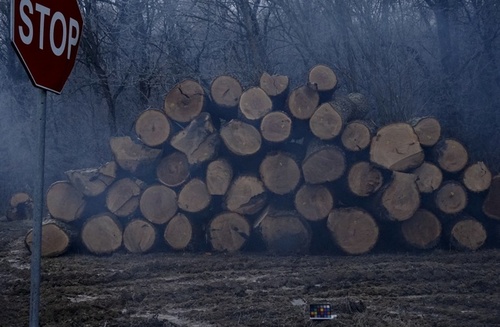}
     \end{subfigure}
     \begin{subfigure}[b]{0.06\textwidth}
         \centering
         \includegraphics[width=\textwidth]{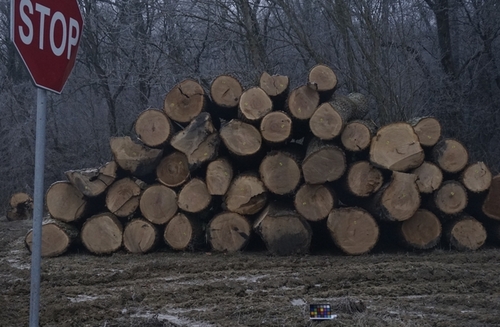}
     \end{subfigure}
     \begin{subfigure}[b]{0.06\textwidth}
         \centering
         \includegraphics[width=\textwidth]{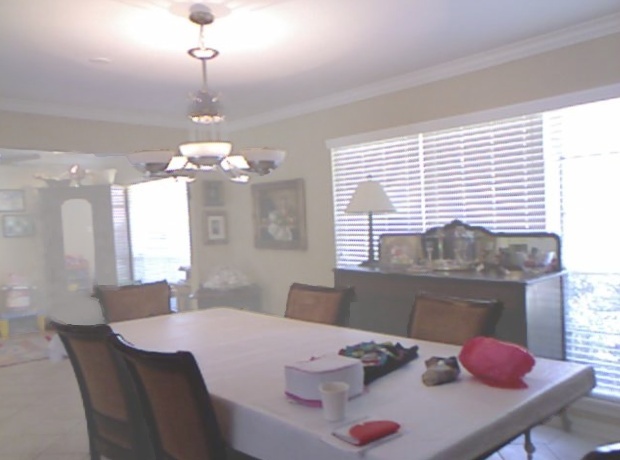}
     \end{subfigure}
     \begin{subfigure}[b]{0.06\textwidth}
         \centering
         \includegraphics[width=\textwidth]{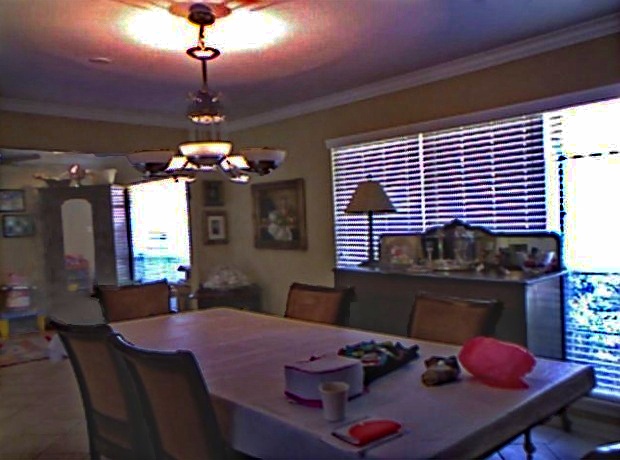}
     \end{subfigure}
     \begin{subfigure}[b]{0.06\textwidth}
         \centering
         \includegraphics[width=\textwidth]{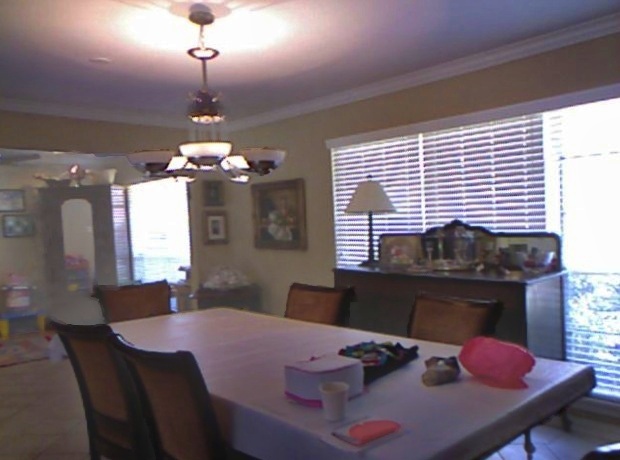}
     \end{subfigure}
     \begin{subfigure}[b]{0.06\textwidth}
         \centering
         \includegraphics[width=\textwidth]{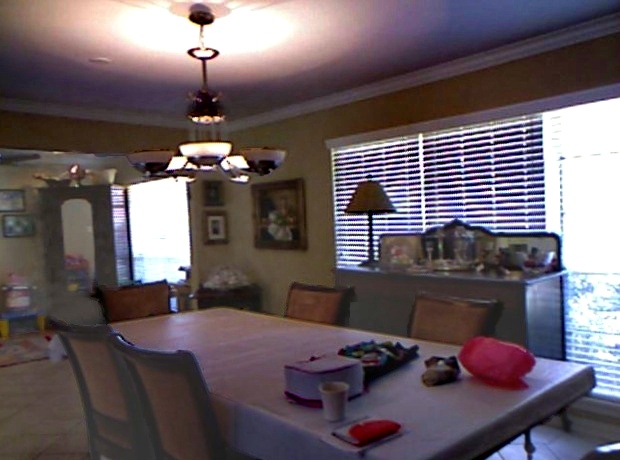}
     \end{subfigure}
     \begin{subfigure}[b]{0.06\textwidth}
         \centering
         \includegraphics[width=\textwidth]{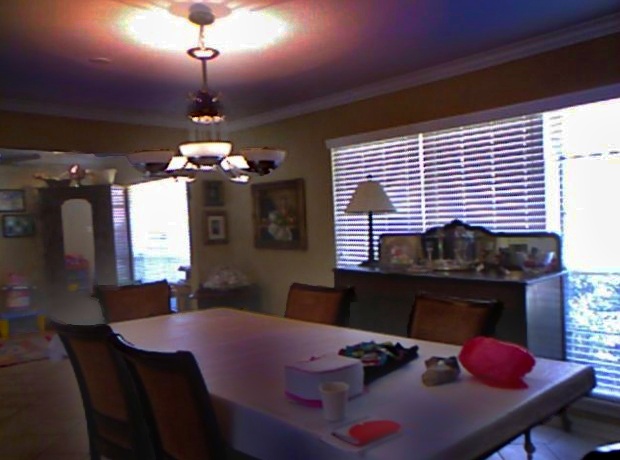}
     \end{subfigure}
     \begin{subfigure}[b]{0.06\textwidth}
         \centering
         \includegraphics[width=\textwidth]{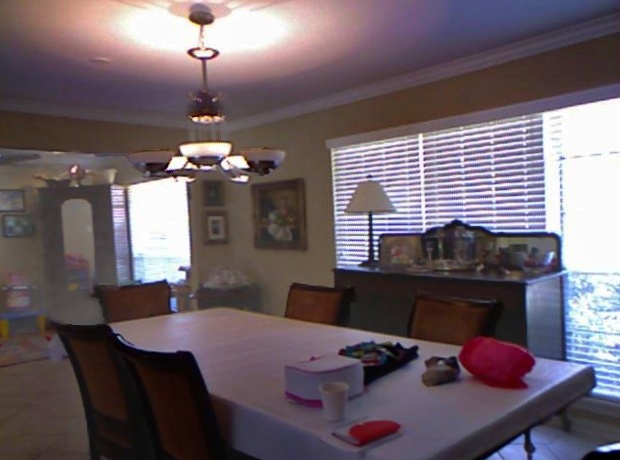}
     \end{subfigure}
     \begin{subfigure}[b]{0.06\textwidth}
         \centering
         \includegraphics[width=\textwidth]{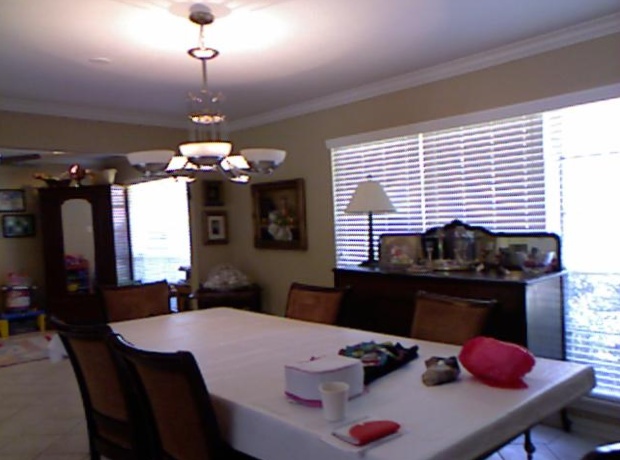}
     \end{subfigure}
     \begin{subfigure}[b]{0.06\textwidth}
         \centering
         \includegraphics[width=\textwidth]{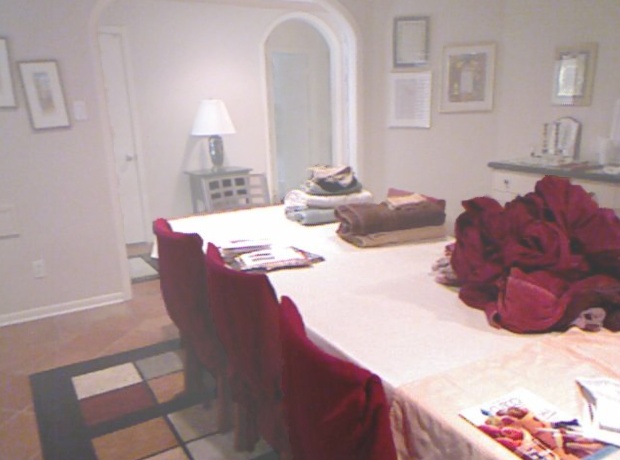}
     \end{subfigure}
     \begin{subfigure}[b]{0.06\textwidth}
         \centering
         \includegraphics[width=\textwidth]{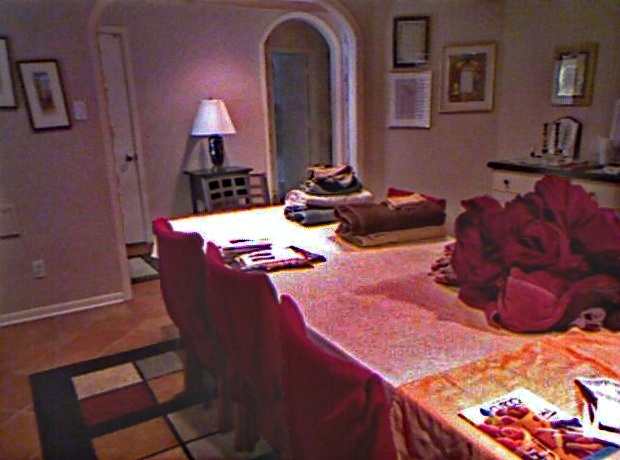}
     \end{subfigure}
     \begin{subfigure}[b]{0.06\textwidth}
         \centering
         \includegraphics[width=\textwidth]{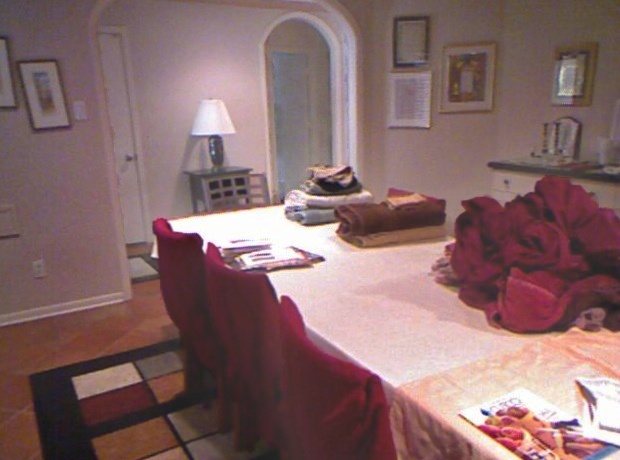}
     \end{subfigure}
     \begin{subfigure}[b]{0.06\textwidth}
         \centering
         \includegraphics[width=\textwidth]{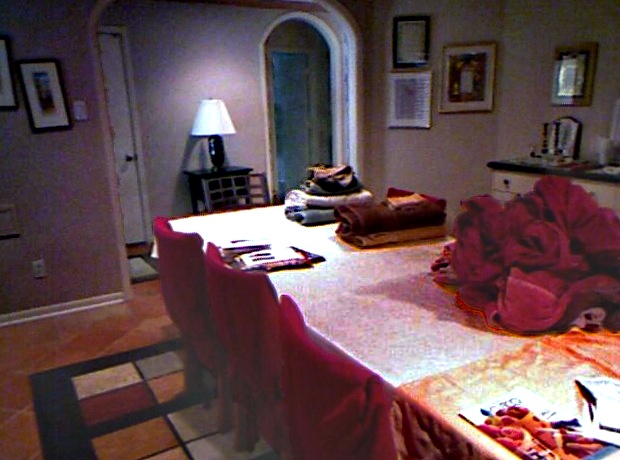}
     \end{subfigure}
     \begin{subfigure}[b]{0.06\textwidth}
         \centering
         \includegraphics[width=\textwidth]{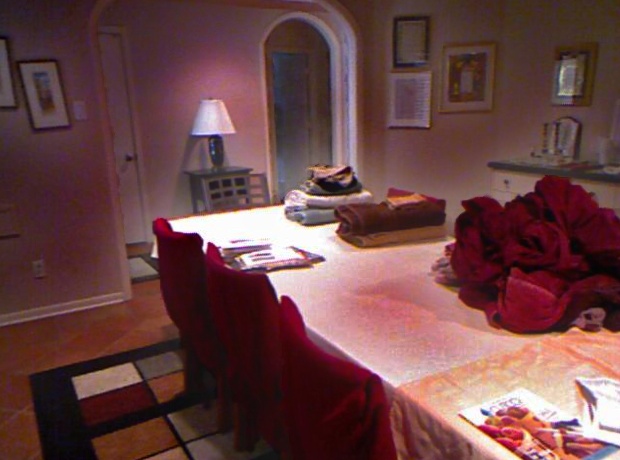}
     \end{subfigure}
     \begin{subfigure}[b]{0.06\textwidth}
         \centering
         \includegraphics[width=\textwidth]{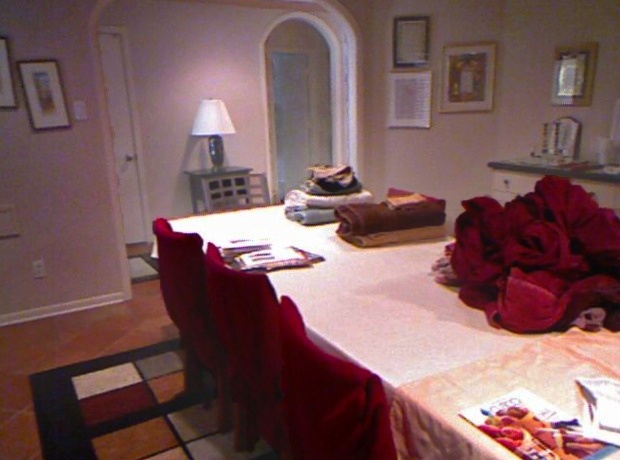}
     \end{subfigure}
     \begin{subfigure}[b]{0.06\textwidth}
         \centering
         \includegraphics[width=\textwidth]{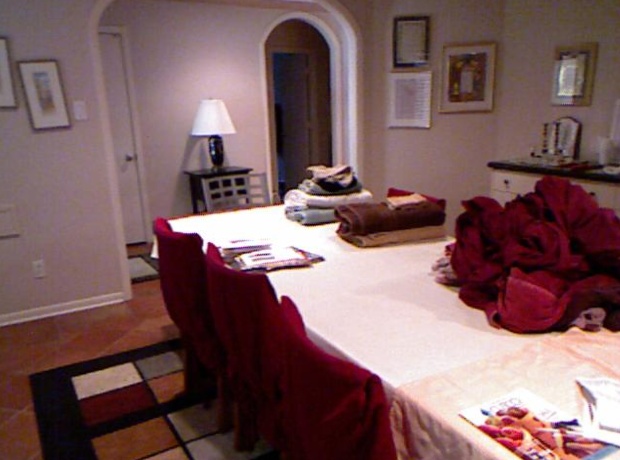}
     \end{subfigure}
     \begin{subfigure}[b]{0.06\textwidth}
         \centering
         \includegraphics[width=\textwidth]{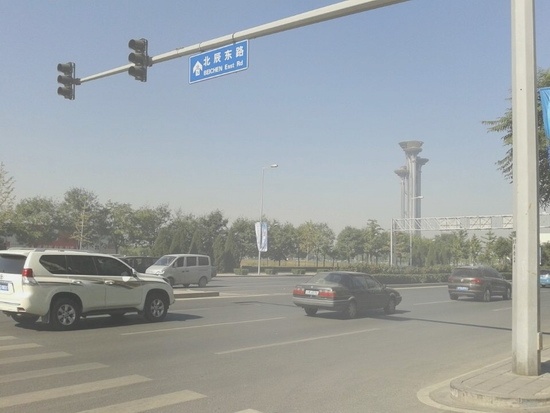}
     \end{subfigure}
     \begin{subfigure}[b]{0.06\textwidth}
         \centering
         \includegraphics[width=\textwidth]{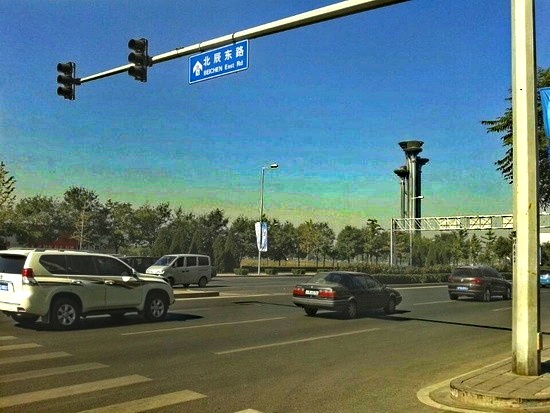}
     \end{subfigure}
     \begin{subfigure}[b]{0.06\textwidth}
         \centering
         \includegraphics[width=\textwidth]{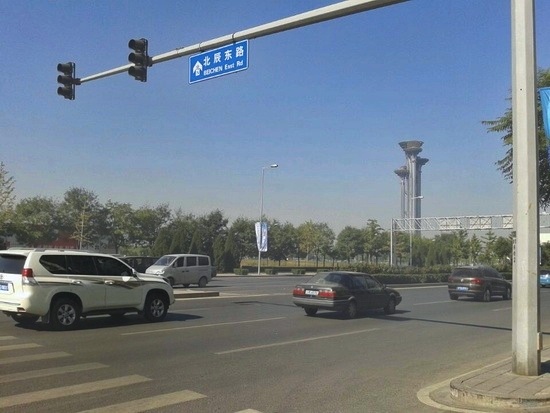}
     \end{subfigure}
     \begin{subfigure}[b]{0.06\textwidth}
         \centering
         \includegraphics[width=\textwidth]{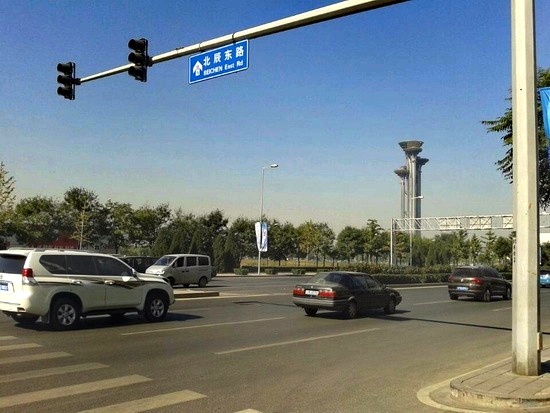}
     \end{subfigure}
     \begin{subfigure}[b]{0.06\textwidth}
         \centering
         \includegraphics[width=\textwidth]{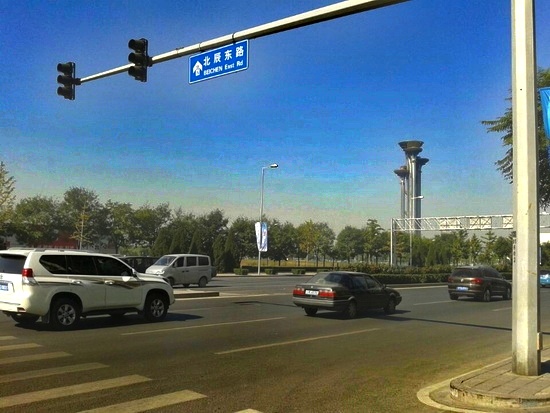}
     \end{subfigure}
     \begin{subfigure}[b]{0.06\textwidth}
         \centering
         \includegraphics[width=\textwidth]{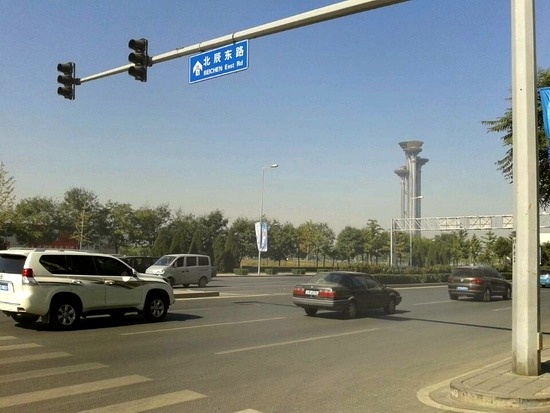}
     \end{subfigure}
     \begin{subfigure}[b]{0.06\textwidth}
         \centering
         \includegraphics[width=\textwidth]{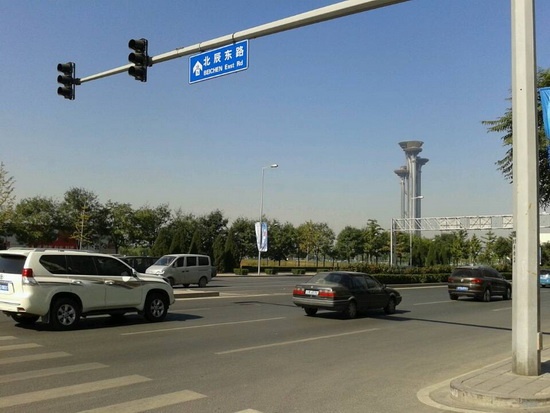}
     \end{subfigure}
     \begin{subfigure}[b]{0.06\textwidth}
         \centering
         \includegraphics[width=\textwidth]{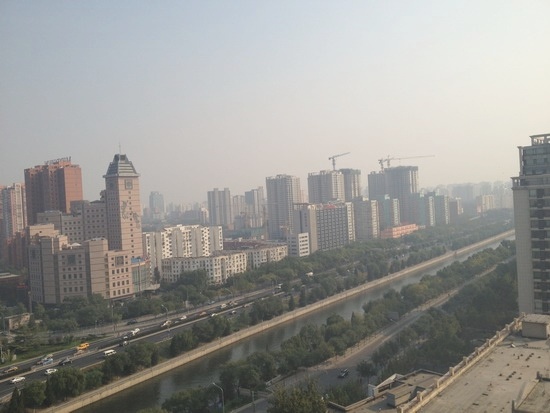}
         \subcaption{}
     \end{subfigure}
     \begin{subfigure}[b]{0.06\textwidth}
         \centering
         \includegraphics[width=\textwidth]{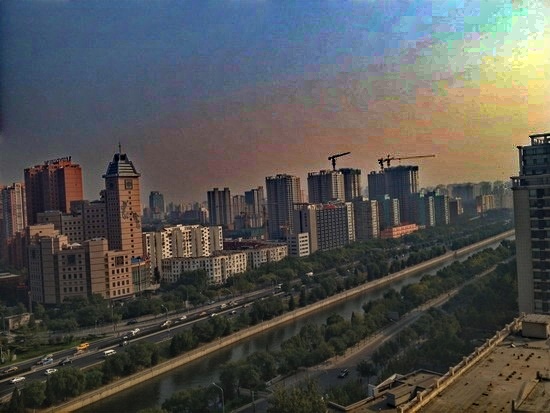}
         \subcaption{}
     \end{subfigure}
     \begin{subfigure}[b]{0.06\textwidth}
         \centering
         \includegraphics[width=\textwidth]{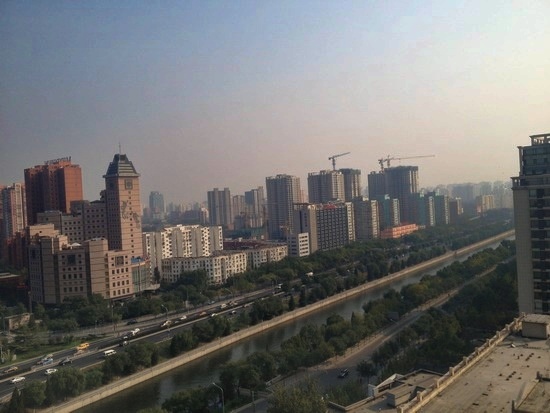}
         \subcaption{}
     \end{subfigure}
     \begin{subfigure}[b]{0.06\textwidth}
         \centering
         \includegraphics[width=\textwidth]{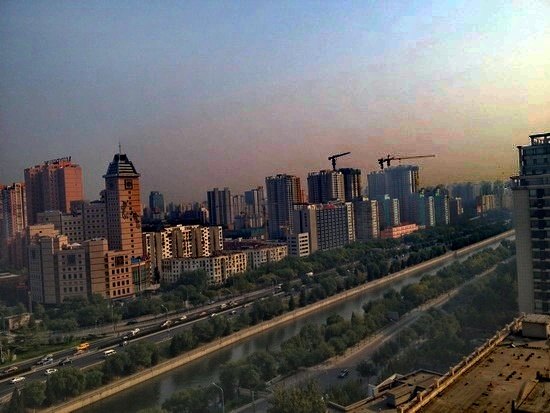}
         \subcaption{}
     \end{subfigure}
     \begin{subfigure}[b]{0.06\textwidth}
         \centering
         \includegraphics[width=\textwidth]{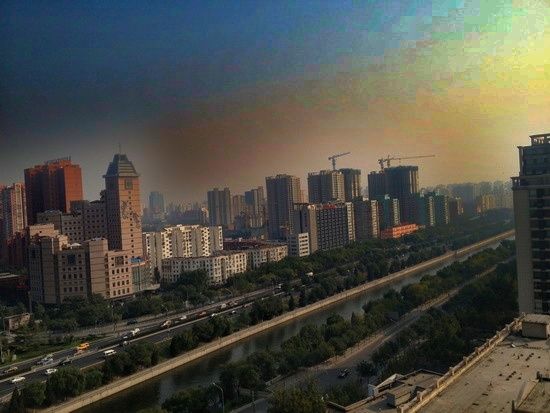}
         \subcaption{}
     \end{subfigure}
     \begin{subfigure}[b]{0.06\textwidth}
         \centering
         \includegraphics[width=\textwidth]{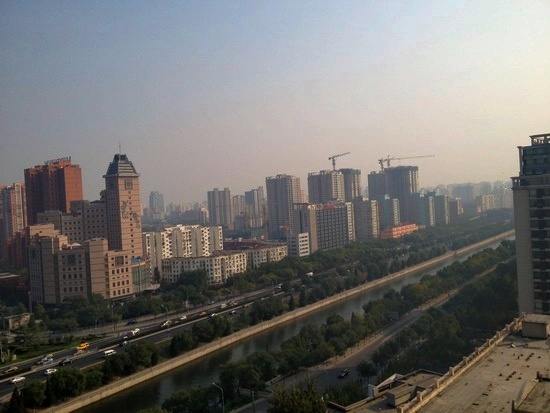}
         \subcaption{}
     \end{subfigure}
     \begin{subfigure}[b]{0.06\textwidth}
         \centering
         \includegraphics[width=\textwidth]{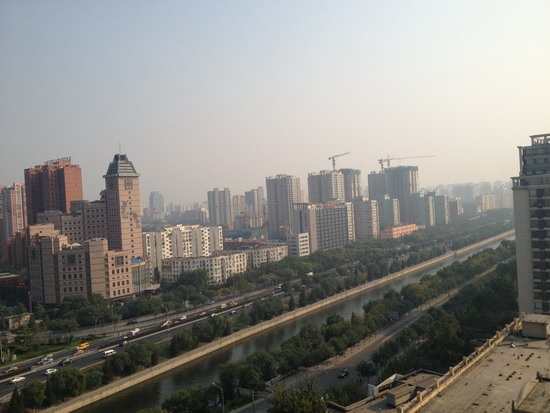}
         \subcaption{}
     \end{subfigure}
     
        \caption{Dehazed images of OHaze (first 2 rows) and RESIDE SOTS (remaining 4 rows). (a) Hazy images. The restored images using (b) BCCR (c) CAP (d) NLD (e) DCP (f) PF-DCP with fusion weight $4:1$(indoor) and $80:1$(outdoor) (g) Ground truth.}
        \label{fig:sots_ohaze}
\end{figure}

As we could see from Fig. \ref{fig:sots_ohaze}, BCCR introduces large color distortion, and the same problem holds for NLD. The results are also reflected in the quantitative analysis. As shown in Table \ref{tab:deep learning}, the PSNR scores for both BCCR and NLD are low due to the inauthentic color. For CAP, the colors are close to those in reality but there is still a lot of haze remaining after dehazing. Thus, CAP achieves a fairly good yet not satisfactory performance in PSNR and SSIM. As for the original DCP, there are apparent halo artifacts around the edges of objects, which comes from the inaccurate transmission rate estimation. Besides, the dehazed images are dark because of the faulty atmospheric light estimation. On the contrary, utilizing the multi-scale information, PF-DCP maintains the authentic color and has little halo artifacts. Therefore, PSNR and SSIM scores for PF-DCP get largely improved compared to the original DCP.

\begin{table}[h!]
  \begin{center}
  \scalebox{0.47}{
    \begin{tabular}{|c|c|c|c|c|c|c|c|c|}
      \textbf{BCCR \cite{meng2013efficient}} &
      \textbf{CAP \cite{zhu2015fast}} &
      \textbf{NLD \cite{berman2016non}} &
      \textbf{AOD \cite{li2017aod}} & \textbf{DCPDN \cite{zhang2018densely}} & \textbf{GFN \cite{ren2018gated}} & \textbf{EPDN \cite{qu2019enhanced}} & \textbf{DCP \cite{he2010single}} & \textbf{PF-DCP} \\
      16.30/0.80 & 20.72/0.87 & 17.68/0.81 &
      19.78/0.87 & 17.98/0.86 & 21.81/0.86 & 22.68/0.90 & 17.82/0.86 & 23.07/0.91 \\
    \end{tabular}
  }
    \caption{Average PSNR and SSIM on the RESIDE SOTS.}
    \label{tab:deep learning}
  \end{center}
\end{table}

In addition to traditional methods, we also perform a quantitative evaluation between PF-DCP with contemporary deep learning methods on RESIDE SOTS. As shown in Table \ref{tab:deep learning}, PF-DCP manages to outperform listed popular deep learning methods in recent years. It is true that the proposed PF-DCP is still inferior to the most recent deep learning methods \cite{shao2020domain}, however, PF-DCP has its own advantages. First, PF-DCP provides reasonably good performance. Second, it does not require computationally intensive model training. Third, it is trackable and explainable rather than a black box. We believe the advantages of PF-DCP reveal great potentials for single image dehazing in real applications.

\section{Conclusion}
In this paper, we have presented an effective algorithm, pyramid fusion dark channel prior (PF-DCP) by introducing multi-scale information of hazy images based on the original DCP. Through experiments, we show that fusing the transmission maps of different image scales can boost the performance of haze removal greatly. Meanwhile, the weighted fusion strategy provides users an alternative to the patch size selection, which is more robust and computationally efficient.
\bibliographystyle{IEEEbib}
\bibliography{bibliography}

\end{document}